\documentclass{article}

\usepackage[numbers, super]{natbib}
\usepackage{threeparttable}
\usepackage{booktabs}

\makeatletter
\renewcommand\@biblabel[1]{#1.}
\makeatother
\usepackage{caption}
\usepackage{float}
\usepackage{graphicx}
\usepackage[utf8]{inputenc} 
\usepackage[T1]{fontenc}    
\usepackage{hyperref}       
\usepackage{url}            
\usepackage{booktabs}       
\usepackage{amsfonts}       
\usepackage{nicefrac}       
\usepackage{microtype}      
\usepackage{xcolor}         
\usepackage{multicol}

\usepackage[final]{neurips_2023}

\title{Enhancing Talent Employment Insights Through Feature Extraction with LLM Finetuning\thanks{This paper was authored in July 2024 for the Applied Analytics Practicum course (CSE/ISYE/MGT 6748) at Georgia Tech (sponsored by AdeptID) and was published on arXiv in January 2025.}}

\author{%
  Karishma Thakrar \\
  Georgia Institute of Technology \\
  \texttt{karishma.thakrar@gatech.edu} \\
  \AND
  Nick Young \\
  Georgia Institute of Technology \\
  \texttt{nyoung@gatech.edu} \\
}

\begin{document}

\maketitle


\begin{abstract}

This paper explores the application of large language models (LLMs) to extract nuanced and complex job features from unstructured job postings. Using a dataset of 1.2 million job postings provided by AdeptID, we developed a robust pipeline to identify and classify variables such as remote work availability, remuneration structures, educational requirements, and work experience preferences. Our methodology combines semantic chunking, retrieval-augmented generation (RAG), and fine-tuning DistilBERT models to overcome the limitations of traditional parsing tools. By leveraging these techniques, we achieved significant improvements in identifying variables often mislabeled or overlooked, such as non-salary-based compensation and inferred remote work categories. We present a comprehensive evaluation of our fine-tuned models and analyze their strengths, limitations, and potential for scaling. This work highlights the promise of LLMs in labor market analytics, providing a foundation for more accurate and actionable insights into job data.

\end{abstract}

\section{Introduction}

AdeptID, a public benefit corporation based in Boston, specializes in advancing talent management, mobility, upskilling, and reskilling through a suite of innovative data products. To support their mission, AdeptID provided access to a database of approximately 1.2 million job postings, enriched with variables (parsed categorizations, entities, and more) to be used as training data.

The primary objective of this practicum was to develop a solution capable of extracting nuanced and machine-interpretable features—or “signals”—from job postings. These signals included critical but often challenging attributes such as remote work availability, salary details, bonuses, and educational or experience requirements. Extracting these features reliably from diverse text-based formats was intended to enable AdeptID to better analyze patterns in their data and provide actionable insights to employers and job seekers.

In keeping with their emphasis on exploration and discovery, AdeptID set intentionally broad requirements, encouraging us to define the scope of the task, decide which features to extract, benchmark performance, and share insights from our findings. While not intended as a production-ready solution, the practicum emphasized research and innovation, with the expectation that our findings would be vetted and integrated with contributions from other teams.

Recognizing the potential of large language models (LLMs) to address the complexities of this task, we leveraged them to develop our solution. We combined data cleaning informed by exploratory data analysis, a representative sample of 10,000 records, and a Gemini LLM to generate initial labels for our target variables. Fine-tuning four separate DistilBERT models for specific features allowed us to focus on nuanced extractions while benefiting from the strengths of pre-trained models. Additionally, we explored advanced techniques like weight-merging to enhance model performance, though our final implementation favored the use of individual fine-tuned models for each variable.

By employing these methods, we successfully developed a robust pipeline for extracting valuable signals from job postings, laying the groundwork for further advancements in labor market analytics.

\section{Methodology}

The methodology employed is outlined in the following flow map. 

\begin{figure}[h]
    \centering
    \includegraphics[width=1\linewidth]{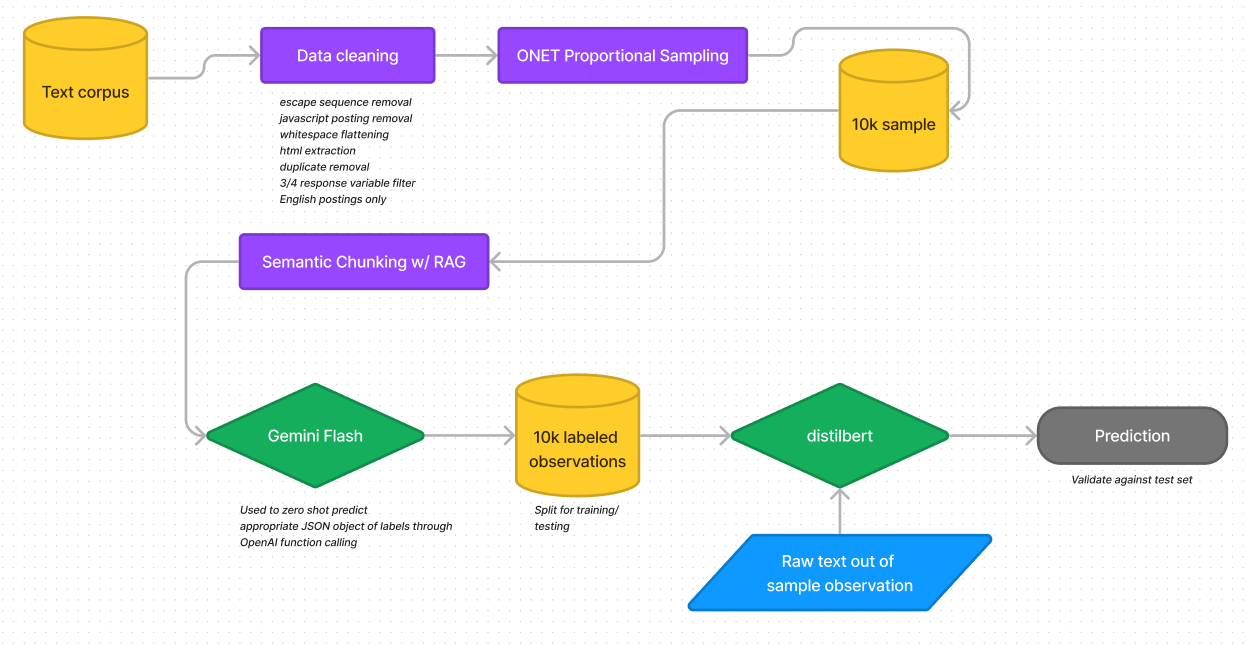}
    \caption{Flow map of our methodology}
    \label{fig:enter-label}
\end{figure}

\subsection{Exploration and Problem-Setting }

Our team performed a comprehensive exploration of the data to identify potential problems to solve for and determine the necessary data cleaning steps. We discovered that certain job features were often inaccurate based on the job posting descriptions.  

To share our findings with AdeptID, we list some high level observations below: 

\begin{itemize}
\item Max education and max experience variables are often NULL, suggesting most jobs don’t have an official cap on education or experience requirements.  
\item Salary-related fields (salary, min salary, max salary) are approximately 50\% NULL. 
\item An estimated 2,000 job postings contain HTML in their raw text, possibly due to traditional parsing service issues. 
\item Most positions (18.7\%)  have the “Unclassified Industry” NAICS 2-digit classification, followed by “Professional, Scientific, and Technical Services” (11.2\%), with “Agriculture, Forestry, Fishing and Hunting” representing the smallest portion of our data (0.17\% ).  
\item Top career areas include Healthcare (13.2\%), IT and computer science (11.0\%), and Finance (7.62\%), while Social Analysis \& Planning, Agriculture and Horticulture, and Personal Services each account for 1.2
\item An almost even breakdown of approximately 1.2\% was observed for the Occupational Information Network (O*NET), a standardized classification system for occupations. 
\item Data Scientists, Information Security, and Software Developers were among the highest paid professions based on salary evaluation by O*NET occupation type.  
\item 31\% of job postings are from California, Texas, Florida, and New York combined. 
\item A significant minority of job postings are internships. This data bias may affect the robustness of our final solution. 
\item Most job postings range from 1,000 to 6,000 characters in length, with some as short as less than 100 and others exceeding 10,000 characters. Not all words are meaningful for feature extraction. 
\item The average character length of a job posting is 4,178 characters, with a standard deviation of 2,746.
\end{itemize}


With a clearer understanding of the data, we were better equipped to ask what types of signals we wanted to uncover. Remembering the emphasis on complex questions with simple answers, decided to focus on the types of information about a job that seem straightforward to us but are actually mired in technical nuance.  

Taking the intersection between our variables of interest and what had been suggested by AdeptID as having more importance, we landed on our final variables being: 

\begin{itemize}
\item Remote work availability 
\item Remuneration types available 
\item Work experience requirements 
\item Educational requirements 
\end{itemize}

At a glance, any of these might seem very straightforward, but their actual state of truth is a bit more complex than what AdeptID typically receives from traditional parsing services. We explain the complications and the approach to these variables discovered through our research below. 

\textbf{Remote type:} The remote work availability variable was maintained in the same format as provided in the database, with job postings classified as either “In Person”, “Remote”, “Hybrid”, or “Null”. In reality, this feature can only take on three real values: at a worker’s location of their choice (remote), a designated workplace (in person), or a combination of both (hybrid). The fourth option of “Null” indicates that the value is unknown or unavailable.  

Our research revealed that traditional parsing solutions often performed poorly in accurately classifying job postings, frequently labeling positions as “Null” despite clear indications of in-person, remote, or hybrid work. In some cases, remote work availability could be inferred from the job’s nature and responsibilities, even if not explicitly stated. 

\textbf{Remuneration:} This variable, inspired by existing salary variables, represents additional information not currently covered. Traditional parsing solutions determine a job’s salary averaging the proposed salary range, multiplying the hourly rate by 40 (hours in a week) and then 52 (weeks in a year), or reading it directly from the job posting. While useful for applicants to know their estimated pay, salary and hourly wages do not cover the full scope of compensation a worker could receive, such as bonuses, commission-based pay or stipends.  

To better expose these nuances, we decided to surface these new variables as boolean indicators determining if the job has salaried or hourly pay and offers bonuses or commission-based pay. These details can help job applicants make more informed decisions on what positions to apply for.  

\textbf{Experience and Education:} The original data includes minimum and maximum years of experience variables but does not indicate if these years are required or preferred. We felt it important to expose to applicants if the requested experience would be a hard limit to their consideration. Thus our feature includes a Boolean variable “experience\_required” and a float value “experience\_minimum\_years” to help applicants know if they need prior experience and the minimum required.  

Similarly for education, we ask if the mentioned education level is required or preferred and include a variable representing the required or preferred education level. This helps applicants understand true qualifications versus preferences.   

\vspace{10pt}

To address these shortcomings, we initially employed regular expressions (regex) and the Natural Language Toolkit (NLTK) to parse job postings for the aforementioned variables, including remote work availability, remuneration, experience, and education. Our very initial goal was to determine the true values of these variables by using regex and NLP approaches to segment the postings and then use these labeled datasets to fine-tune an LLM for improved classification accuracy. We sought to achieve this by searching specific keywords or phrases indicating certain attributes. However, we encountered challenges such as overfitting to specific examples and use cases, resulting in many false positives and negatives, and the need for a more flexible solution. Additionally, given the project’s time constraints, refining and optimizing our regex and NLP approaches would have been time-consuming. These challenges and realizations prompted us to pivot our approach and explore the use of LLMs for classification as a more adaptable solution that could potentially overcome the limitations of our initial methods and deliver accurate classifications within our project timeline.  

Overall, given a job posting, we sought to output the following information:  

\begin{verbatim}
{
"remote_type": string,
"remuneration": {
"is_salaried": boolean,
"is_hourly": boolean,
"has_bonus": boolean,
"has_commission": boolean
},
"experience": {
"experience_required": boolean,
"experience_minimum_years": float
},
"education": {
"requirement_level": string,
"education_level": string
}
}
\end{verbatim}

\subsection{Methodology: Approach to the Larger LLM}

Our approach to solving the problem began with data cleaning based on our previous EDA findings. This include: 

\begin{itemize}
    \item Removing extra whitespace 
    \item Removing escape character such as "\textbackslash n" and "\textbackslash t"
    \item Parsing raw text from HTML embedded text 
    \item Removing postings that we could not successfully parse raw text from JavaScript 
    \item Removing duplicative postings 
    \item Removing job postings that had NULL values for 3 of 4 of our variables of interest 
    \item Removing job postings that were not in English 
\end{itemize}

To ensure our small sample was representative of the population, we took a weighted sample by O*NET. This was done because our EDA revealed that certain job postings were more likely to follow either structured or unstructured formats. By using a weighted sample with the same O*NET breakdown as the total dataset, we ensured our model would be trained on a diverse range of data. 

After evaluating several LLMs, we chose Gemini 1.5 Flash, a large commercial LLM, to zero-shot predict the labels of our new features for a sample of roughly 10,000 observations. Gemini was selected for its one million token context window and a starting price that was a tenth of the cost of other state-of-the-art models, as determined from our cost-benefit analysis. Using these labels as ground truth, we planned to fine-tune a pipeline of smaller open-source LLMs to serve as the final layer, delivering predictions for the variables in JSON format when given a job description. 

We faced several considerations, including determining the relevant information to pass to the LLM, ensuring consistent response formats without hallucinating unacceptable values, and working within our available computational resources. Given time and resource constraints, we proceeded with faith in our chosen techniques rather than following the typical iterative data science lifecycle. 

To provide our LLM with the most relevant information and limit the token size, we implemented semantic chunking and retrieval-augmented generation (RAG) in our pipeline \cite{nayak2024}. This approach breaks the text corpus into smaller chunks and retrieves the most relevant chunks based on the query. 

We used the TokenTextSplitter to split the text into meaningful chunks based on the semantic structure of the text, ensuring that the chunks maintain context and relevance. We chose the JINA Ai embedding model from HuggingFace \cite{jina_embeddings} for vector-enabled storage and retrieval due to its high ranking, reasonable parameter training, consideration of hard negatives, and suitability for fast inference on larger documents. The FAISS vector store is employed to store the embeddings of chunks from individual job postings, allowing for efficient retrieval of the most relevant chunks based on a given query. 

Vector storing and retrieval offers several advantages, such as overcoming the limitations of a model's context window, providing more accurate responses, and improving overall pipeline performance. These techniques form the basis of our path from job posting to LLM query. Given a job posting, we semantically chunk the text and impart vector embeddings. When the LLM is passed a prompt, the vector storage is leveraged to only return the chunks that are most similar to the prompt, thereby bolstering relevancy in a lesser amount of tokens. 

After addressing the input side of our first LLM, we needed to implement controls for the output format. We researched packages and techniques such as jsonformer \cite{jsonformer}, OpenAI function calling \cite{greyling2023}, and PyDantic \cite{chinnock2024}, but ultimately found it most effective to include specific formatting instructions and acceptable values within our prompt. This approach allowed us to constrain the LLM's output to a desired format and range of values. 

However, even after completing our implementation, we encountered inconsistent and redundant outputs. For example, terms like “on\-site," “on\_site," and “in\_person" all referred to the same concept in the remote\_type variable. To ensure data consistency, we mapped the unique outputs to a defined number of output classes for each of our variables of interest. This process resulted in a significant reduction of output classes for some variables, such as an 80\% reduction for education. 

The remaining decisions involved refining the prompts and certain parameters of our chunking implementation. We found that using a separate prompt and LLM for each of our four variables of interest yielded the most precise results. We iteratively tuned these prompts by generating output for 50 samples with hand-labeled true values and making adjustments that improved the accuracy of the LLM pipelines across those samples. In total, we tested over 100 prompts with slight variations during this process. 

\subsection{Approach to the Smaller LLM}

Having four separate chunking/prompt versions for the four different variables, we ran these across our nearly 10,000 records to create a training dataset that we assumed would represent the ground truth based on the performance we observed on the 50 hand-annotated samples. We then created four separate fine-tuning pipelines using the DistilBERT model as our starting point LLM.  

We chose DistilBERT from the models available on HuggingFace because of its versatility across a wide range of use cases and its popularity within the community, as evidenced by its many recent downloads. It's a compact and more efficient version of the BERT model, created using knowledge distillation. This technique allows DistilBERT to be 40\% smaller than BERT while maintaining 97\% of its language understanding abilities \cite{distilbert}. Its small size and strong performance made it an ideal fit for our requirements. 

In total, we initiated four separate fine-tuning blocks for our four different variables with each of these fine-tuning processes taking between 4 and 9 hours to complete across both our PCs. Initially, when building this pipeline, we tested the AutoTokenizer but ultimately chose the DistilBertTokenizer as it was optimized for our fine-tuned model and demonstrated better performance based on our tests. 

In an effort to enhance the model's performance, we investigated weight merging techniques to consolidate the four fine-tuned DistilBERT models into a single, unified model. To implement this, we developed a custom model class that leveraged a shared DistilBERT base and employed a single classifier head with an expanded number of output labels. We also created a function that combined the weights of the individual models' classifier layers into the combined model's classifier layer. 

We anticipated that this innovative weight-fusion strategy would yield superior results due to the potential for knowledge sharing and generalization across the various classification tasks, resulting in positive task transfer \cite{savvov2024}. However, upon evaluation, the performance of the merged model was found to be subpar compared to the use of separate fine-tuned models. As a result, we decided to proceed with the approach of combining the outputs of the individual fine-tuned models without weight merging to generate the final predictions, as this method demonstrated better performance in our experiments. 

In summary, the final pipeline takes a job description as input, performs semantic chunking and retrieval which are then passed through the fine-tuned DistilBERT models. The final output is a dictionary containing the predicted labels and their corresponding scores for each of the four variables, allowing us to extract the relevant information from the job description in a structured format.  

\section{Evaluation}

For each fine-tuning block, we implemented an 80/20 train/test split to get a baseline assessment of how our models performed. Here were our results:

\begin{table}[h]
\begin{center}
\caption{Evaluation metrics per variable (abbreviated)}

\bigskip

\begin{tabular}{|l|llll||}
\hline
Metric & Rmt & Rmn & Exp & Edu \\ \hline
F1 & 0.78 & 0.68 & 0.59 & 0.66 \\
Precision & 0.77 & 0.69 & 0.58 & 0.66 \\
Recall & 0.78 & 0.70 & 0.60 & 0.67 \\
MCC & 0.60 & 0.63 & 0.53 & 0.59\\
\hline

\end{tabular}
\end{center}
\end{table}

To optimize our resources and time, we treated the entire JSON dictionary of each feature as a single label, enabling us to predict 9 distinct variables using only 4 LLM pipelines. This decision was driven by computational and time constraints, as fine-tuning a single LLM often took between 4 and 9 hours. The evaluation metrics presented are based on the entire JSON dictionary label, with the exception of the remote\_type variable, which only has one value in its dictionary. This means that for a prediction to be considered accurate, the LLM needed to correctly predict all sub-variables within the dictionary. 

We did also unnest the actual and predicted JSON dictionaries into their own variables to better evaluate the performance of our method. These results are shown below. 

\begin{table}[h]
\begin{center}
\caption{Accuracy per sub-variable}

\bigskip

\begin{tabular}{l|l}
\hline
Sub-variable & \% correct \\ \hline
Remote Type & 77.4 \\
Is Salaried & 85.8 \\
Is Hourly & 86.0 \\
Has Commission & 96.4 \\
Has Bonus & 84.4 \\
Experience Required & 74.2 \\
Experience Level & 64.3 \\
Education Required & 73.5 \\
Education Level & 85.6 \\
\hline

\end{tabular}
\end{center}
\end{table}

It's important to acknowledge that the ground truth for the features in our test set was based on the output of a larger LLM. Nevertheless, our smaller LLM demonstrated a reasonable ability to be fine-tuned and follow some of the same logic as the larger LLM. When it comes to the education variable, the prediction of the required level performed better than the prediction of whether that level was required. Conversely, the opposite is true for experience. We believe this could be attributed to how these levels of requirement are expressed within job postings. Educational level is typically conveyed using specific and standardized words (Bachelor's, Masters, PhD, etc.), while experiential requirement is expressed using an integer of years, which might appear frequently throughout the job posting. And so, the association between requirement/preference and education might be easier for the LLM to interpret compared to the same association with years of experience. 

For the remuneration variable, the commission feature performs the best, but this could be due to a highly skewed class distribution. Most jobs typically do not offer commission-based pay. This might be evident in our training data and, consequently, emulated in our test data. However, this does not guarantee that the model will perform well in instances where commission-based pay is present; the model might be guessing that most jobs don't have a commission, which is generally true. 

To further assess the usefulness of our model, we evaluated its performance on a small sample of 10 job postings. This sample consisted of 5 real job postings from the database not used in training or testing, and 5 job postings we created to represent a more human-language form with some general observations below: 

\begin{itemize}
\item Only 4 out of the 10 records were correct across all variables, 3 of which were from our handmade records, which might have been easier for the LLM to understand due to their concise nature. 
\item Only 1 record had errors across more than 1 variable category (remote type and remuneration). 
\item The most mislabeled variable was education, which was incorrect in 3 records, followed by experience, with 2 mislabeled records. 
\end{itemize}

In addition to the model's performance, it's important to consider the cost of this pipeline when assessing its usefulness. Making an inference from a job posting incurs no cost, as the smaller LLMs are open-source. Moreover, generating approximately 10,000 record ground-truth labels through Gemini incurred minimal charges overall. On average, making an inference from the final pipeline takes 1.1 seconds. It's also worth noting that generating the training data for each of the 4 individual variable pipelines took between 8 to 14 hours, and fine-tuning each pipeline took an additional 4 to 9 hours, resulting in an approximate total of +80 hours of building to using our pipeline. 

\section{Next Steps \& Recommendations}

While we were unable to attempt all the techniques we wanted to try, we believe some of these could be potentially be improved with some refactoring. We also believe we can employ simpler solutions like regex for preprocessing and later in our pipeline. Regex could be a practical alternative to semantic chunking and RAG for improving performance. The concept behind RAG with chunking is to provide the LLM with the most relevant parts of a corpus to answer a question, by splitting the corpus into documents and retrieving a few based on their similarity to the prompt. We heard from other teams that they used regex to identify keywords and extract relevant sections of the corpus for their LLMs as our method of RAG. Given more time, we could have applied this approach using keywords related to remote work, remuneration, experience, and education to selectively extract pertinent sentences for our LLMs. 

In addition to regex, we could have explored other preprocessing techniques such as stop word removal and lemmatization. Stop word removal involves eliminating common words that do not contribute significantly to the meaning of the text, such as “the,” “a,” and “an.” This can help reduce noise in the data and improve the efficiency of the LLM. Lemmatization, on the other hand, involves reducing words to their base or dictionary form, which can help normalize the text and reduce the dimensionality of the input data too.  

We also observed that LLMs sometimes confused negated requirements, such as interpreting “you do not need a Bachelor's degree" as the job requiring a degree. Handling negation in NLP tasks is a nuanced topic. One potential solution is to concatenate negating words with the following word (e.g., “notneed" instead of “not need") to create a tokenizable difference that the LLM can better understand. 

Moreover, if we had more time and resources, we might consider using one smaller LLM for each of the nine sub-variables, instead of four larger LLMs handling remote type, remuneration, education, and experience. We hypothesize that this approach could allow each LLM to better understand the patterns in the training data when the labels consist solely of one sub-variable at a time. We could have also explored more embeddings models and open-source LLMs when finetuning our data. 

Finally, with further research, we could have explored additional forms of weight merging  that were more performant than what we saw in our practice. This could involve investigating different techniques for combining the outputs of multiple LLMs, such as weight averaging, majority voting, or more advanced ensemble methods. By leveraging the strengths of various weight merging approaches, in addition to testing our other areas of interest, we might be able to improve the overall performance and robustness of our pipeline further.  

\section{Conclusion}

In this work, we developed a pipeline leveraging large language models (LLMs) to extract nuanced and machine-interpretable features from job postings, addressing the limitations of traditional parsing tools. By combining exploratory data analysis, semantic chunking, RAG, and fine-tuning DistilBERT models, we demonstrated the feasibility of accurately identifying critical variables such as remote work availability, remuneration types, experience requirements, and educational qualifications. Our approach highlights the potential of LLMs to provide actionable insights for labor market analytics, despite the inherent complexity of the data.

While our models achieved reasonable performance, challenges such as data inconsistencies, nuanced language patterns, and computational constraints limited the overall accuracy of our predictions. These limitations underscore the need for further exploration of preprocessing techniques, more granular model architectures, and advanced ensemble methods to enhance performance. Despite these challenges, our solution offers a scalable framework that can be refined and extended to better meet the needs of stakeholders like AdeptID.

We greatly appreciate this opportunity to partner with AdeptID as part of our Applied Analytics Practicum at Georgia Tech. This collaboration highlights the transformative potential of LLMs in addressing complex, real-world challenges. By bridging gaps in labor market analytics, this project provides a foundation for empowering stakeholders with richer, more reliable insights to support informed decision-making in talent mobility, upskilling, and reskilling efforts. Looking ahead, future work will focus on refining our methodology, broadening the range of extracted features, and exploring cost-effective solutions to further enhance the pipeline’s accuracy and resilience.

All team members have contributed a similar amount of effort.

\newpage
\setcounter{footnote}{1}
\bibliography{mybib} 

\end{document}